\pdfoutput=1

\documentclass[11pt]{article}

\usepackage[]{acl}

\usepackage{times}
\usepackage{latexsym}
\usepackage{epigraph}
\usepackage{tabu}
\usepackage{booktabs}
\usepackage[shortlabels]{enumitem}
\usepackage{mathrsfs}

\usepackage[T1]{fontenc}
\DeclareUnicodeCharacter{0307}{\.}

\usepackage[utf8]{inputenc}

\usepackage{microtype}
\usepackage{graphicx}
\usepackage{wrapfig}
\usepackage{enumitem}
\usepackage[font=small]{caption}
\usepackage{amsmath}
\usepackage{amssymb}
\usepackage{cleveref}
\usepackage{bbm}
\usepackage{dsfont}
\DeclareMathOperator*{\argmax}{argmax}
\DeclareMathOperator*{\argmin}{argmin}

\title{It's MBR All the Way Down: \\ Modern Generation Techniques Through the Lens of Minimum Bayes Risk}

\author{Amanda Bertsch\thanks{\hspace{1.5mm}Denotes equal contribution.} \and
  Alex Xie$^*$ \and 
  Graham Neubig \and 
  Matthew R. Gormley\\
    Carnegie Mellon University \\
    \texttt{[abertsch, alexx]@cs.cmu.edu} \\
  }

\begin{document}
\maketitle
\begin{abstract}
Minimum Bayes Risk (MBR) decoding is a method for choosing the outputs of a machine learning system based not on the output with the highest probability, but the output with the lowest risk (expected error) among multiple candidates. It is a simple but powerful method: for an additional cost at inference time, MBR provides reliable several-point improvements across metrics for a wide variety of tasks without any additional data or training. Despite this, MBR is not frequently applied in NLP works, and knowledge of the method itself is limited. We first provide an introduction to the method and the recent literature. We show that several recent methods that do not reference MBR can be written as special cases of MBR; this reformulation provides additional theoretical justification for the performance of these methods, explaining some results that were previously only empirical. We provide theoretical and empirical results about the effectiveness of various MBR variants and make concrete recommendations for the application of MBR in NLP models, including future directions in this area.
\end{abstract}

\section{Introduction}
\epigraph{“Sometimes innovation is only old ideas reappearing in new guises … [b]ut the new costumes are better made, of better materials, as well as more becoming: so research is not so much going round in circles as ascending a spiral.”}{\cite{Jones1994}}

Minimum Bayes Risk (MBR) decoding (\citet{bickel-doksum-1977}; \S \ref{sec:background}) is a decoding method following a simple intuition: when choosing a best output from a set of candidates, the desirable output should be both 1) high probability and 2) relatively consistent with the rest of the outputs (i.e., outputs that are not consistent with the other outputs are high \textit{risk}-- they may be dramatically better or worse than the consensus). 
MBR thus provides an alternative to the more standard maximum-likelihood decoding; when a sample of sufficient size is taken, MBR almost uniformly outperforms beam search and single-output sampling across tasks, metrics, and datasets (see \S \ref{sec:mbrnlp}). 
It is also notable in its flexibility; in \S \ref{sec:best-practices} we organize and discuss several different design decisions that go into the use of MBR and how they affect the efficacy of the method.

While MBR is rarely applied by name in modern NLP, a number of methods with similar intuitions have gained popularity. In \S \ref{sec:itsallmbr}, we demonstrate that a number of generation techniques widely used with modern language models can be viewed as special instances of MBR: 
    \textbf{self-consistency} \cite{wang2023selfconsistency} and its extensions, \textbf{range voting} \cite{borgeaud-emerson-2020-leveraging}, \textbf{output ensembling} \cite{denero-etal-2010-model,martinez-lorenzo-etal-2023-amrs}, and some types of \textbf{density estimation} \cite{kobayashi-2018-frustratingly}. This view exposes connections between seemingly disparate methods and presents theoretical justifications for existing empirical results using these methods. We also discuss how insights from the MBR literature can inform the use of these other MBR-like methods. 

    With the framing of MBR, the theoretical justification for the empirical performance of several methods becomes clear; the extension of self-consistency to open-ended generations becomes trivial; and several promising modifications to self-consistency and output ensembling are exposed.  
    In particular, modern MBR-like methods often do not apply the insights from research on MBR,
    suggesting that these methods could be further improved. 
    In \S \ref{sec:experiments}, we show that some design choices, though seemingly intuitive to a practitioner accustomed to search-based decoding methods, should be avoided when applying MBR.

\section{Formalization}
\label{sec:background}
We begin with the basics of decoding and MBR.
\subsection{Standard decoding}

Decoding from an autoregressive model (such as a transformer decoder) is performed tokenwise. The distribution at each decoding step is conditioned on the prior tokens and the input text:
\begin{align}
p(y_i|y_{<i}, x)
\end{align}
The model is \textit{locally normalized}; the probabilities of next tokens sum to 1.
The probability of a sequence under this global model distribution is 
\vspace{-0.2cm}
\begin{align}
p(y|x) = \prod_{i=1}^{T}p(y_i|y_{<i}, x)
\label{eq:modeldist}
\end{align}

Given this distribution, there are several ways of extracting an output: by sampling at each decoding step from the distribution over next tokens (often with some modification to the distribution, e.g. temperature, nucleus, or epsilon sampling; \citet{holtzman2019curious}); by always choosing the most probable next token (i.e. greedy decoding); or by performing a search over some subset of the output space, guided by the distribution (e.g. beam search, best-first search). These methods generally return a single output; if multiple output candidates are present, the one with the \textit{maximum likelihood} under the model distribution is returned.

\subsection{Minimum Bayes Risk decoding}
\label{sec:mbrmath}

The traditional formulation of MBR is as a minimization objective. Given a \textit{output space} $\mathscr{Y}$ and a probability distribution over this space $p(y|x)$, we compute the risk $R(y')$ of a candidate decoding $y'$ as the expected error (also called \textit{loss}) under this distribution \cite{bickel-doksum-1977, kumar-byrne-2004-minimum, tromble-etal-2008-lattice}. The MBR decoding is then the $y'$ within $\mathscr{Y}$ that minimizes risk:
\vspace{-0.5cm}
\begin{align}
\label{eq:mbr-risk}
\hat{y} &= \argmin_{y' \in \mathscr{Y}} R(y') \\
&= \argmin_{y' \in \mathscr{Y}} \mathbb{E}_{y \mid x}[L(y, y')] \\
&= \argmin_{y' \in \mathscr{Y}} \sum_{y \in \mathscr{Y}} L(y, y') p(y | x)
\end{align}

We can trivially rewrite the risk as a maximization of gain (also called \textit{utility}) rather than a minimization of error, where $G(y,y') = -L(y,y')$.

\paragraph{Approximating risk} Computing this sum over the space of all possible outputs $\mathscr{Y}$ is intractable for most models.\footnote{This is the case for many deep generative models, such as a transformer language model and other autoregressive models without conditional independence assumptions.} In these cases, we approximate the risk $R(y')$ by using a subset of the full space $\mathcal{Y} \subset \mathscr{Y}$
; that is, instead of exact computation of the expectation, we approximate it with a sum over independent samples from $p(y|x)$. Generally, this is performed by sampling repeatedly from a model (or several models) and estimating the probability of each individual output as proportional to the relative frequency that the output occurs.\footnote{This is called a Monte Carlo approximation.} For an unbiased sampling method\footnote{We discuss the use of biased samplers in \S \ref{sec:sample-evidence} and \S \ref{sec:sample-hypothesis}.} (e.g. ancestral sampling), as the number of outputs drawn goes to infinity, this recovers the model's true distribution of probability over sequences. Thus, we approximate risk using this sample: 
\vspace{-0.3cm}
\begin{align}
\label{eq:mbr-mc}
R(y') \approx  \frac{1}{|\mathcal{Y}|} \sum_{y \in \mathcal{Y}} L(y,y') \\ = - \frac{1}{|\mathcal{Y}|} \sum_{y \in \mathcal{Y}} G(y,y')
\end{align}

 Thus, given a sample (which may include duplicates) $\mathcal{Y}$ and a function (e.g. a metric) that compares two sequences $G: \mathscr{Y} \times \mathscr{Y} \rightarrow \mathbb{R}$, we approximate the true MBR decoding rule as:
\vspace{-0.2cm}

\begin{equation}
\label{eq:mbr-mc-gain}
\hat{y} =  \argmax_{y' \in \mathcal{Y}} \frac{1}{|\mathcal{Y}|} \sum_{y \in \mathcal{Y}_e} G(y,y')
\end{equation}

\paragraph{Separation of evidence and hypothesis sets}
\label{sec:sep-set}
In many cases, the same subset of the output space is used for both the risk estimate and the candidate outputs. However, when the sample is substantially smaller than the full output space, it is often beneficial to use separate sets \cite{eikema-aziz-2022-sampling,yan2023dcmbr}. Following prior work (\S \ref{sec:sep-set}), we refer to these as the \emph{evidence set} ($\mathcal{Y}_e$) and \emph{hypothesis set} ($\mathcal{Y}_h$). 

This separation is beneficial because there are distinct and potentially contradictory desiderata for the two sets. We wish for our evidence set to cover a large, representative portion of the search space to obtain a more accurate estimate of risk. However, we want our hypothesis set to only cover the narrower, high-quality region of the space, as we do not want to consider candidate hypotheses that are low-quality. Applying the separation of evidence and hypothesis sets yields the equation for MBR over two subsets of the output space:

\begin{equation}
\label{eq:mbr-gain}
\hat{y} =  \argmax_{y' \in \mathcal{Y}_h} \sum_{y \in \mathcal{Y}_e} G(y,y')
\end{equation}

\section{Taxonomy of MBR}
\label{sec:best-practices}
Equation \ref{eq:mbr-gain} demonstrates four major axes along which an MBR method may vary:
\begin{enumerate}    \setlength{\itemsep}{0.5pt}
    \item Choice of hypothesis set $\mathcal{Y}_h$
    \item Choice of evidence set $\mathcal{Y}_e$
    \item Choice of gain (or error) function $G(y, y')$
    \item Choice of evidence distribution $p(y|x)$
\end{enumerate}
In this section, we examine how these four factors affect the efficacy of MBR and give recommendations for each; in Section~\ref{sec:itsallmbr}, we discuss how these apply to other MBR-like methods.

\begin{table*}[t]
    \centering
    \resizebox{\textwidth}{!}{
    \begin{tabular}{l l l l l l l }
    
        \toprule

        {\textbf{Method}} & Evidence Gen. & Hypothesis Gen. & Metric & $p(y|x)$  \\

        \midrule
        Lattice MBR \cite{tromble-etal-2008-lattice} & N-best list & N-best list & BLEU & translation lattice \\
        Coarse-to-fine MBR \cite{eikema-aziz-2022-sampling} & ancestral sampling & $\texttt{filter}$(sample) & \textsc{BEER} & single model \\ 
        \citet{wiher-etal-2022-decoding} &  ancestral sampling & evidence + more decodings & \textsc{BEER} & single model \\
        MBR-DC \cite{yan2023dcmbr} & temperature sampling\footnotemark[1] & temperature sampling\footnotemark[1] & BLEURT & single model \\
        Ours (\S~\ref{sec:whichmetric}) & ancestral sampling & temperature sampling & BERTScore &  single model \\
        Ours (\S~\ref{sec:whichprob}) & ancestral sampling & temperature sampling & BERTScore &  length-corrected scores \\
        \cite{freitag2023epsilon} &  \multicolumn{2}{c}{epsilon sampling} & \textsc{BLEURT} & single model \\
        Crowd sampling\footnotemark[2] \cite{suzgun-etal-2023-follow} &  \multicolumn{2}{c}{temperature sampling} & neural score metric & single model \\ 
        MBR-Exec \cite{shi-etal-2022-natural} & \multicolumn{2}{c}{temperature sampling} & execution match & single model \\

        \midrule

        Self-consistency (SC) \cite{wang2023selfconsistency} & \multicolumn{2}{c}{temperature sampling} & exact answer match &  single model \\

        Complex SC  \cite{Fu2022ComplexityBasedPF} & \multicolumn{2}{c}{$\texttt{filter}$(temperature sample)} & exact answer match &  single model\\

         SC  for open-ended gen \cite{jain2023selfconsistency} & \multicolumn{2}{c}{temperature sampling} & n-gram overlap &  single model\\
        Range voting \cite{borgeaud-emerson-2020-leveraging} & \multicolumn{2}{c}{beam search} & n-gram overlap &  single model\\

        Post-Ensemble \cite{kobayashi-2018-frustratingly} & \multicolumn{2}{c}{beam search for each model in ensemble} & cosine similarity & model set \\

        AMRs Assemble! \cite{martinez-lorenzo-etal-2023-amrs} & model set & beam search & perplexity & model set \\
        \bottomrule
        
    \end{tabular}
    }
    \caption{Recent work under our taxonomy.
    The line separates methods that are explicitly MBR (above) from those that we identify as MBR-like (below).
    \\ 
    \footnotemark[1] Different temperatures used for evidence and hypothesis.\\
    \footnotemark[2] While \citet{suzgun-etal-2023-follow} coin the new term \textit{crowd sampling}, they also explicitly refer to their method as MBR.
    }
    \label{tab:mbrtaxonomy}
\end{table*}

\subsection{Sampling a hypothesis set}
\label{sec:sample-hypothesis}

Several recent works show benefits from improving the quality of the hypothesis space.
 \citet{fernandes-etal-2022-quality} apply a two-stage approach where they first apply an $N$-best (referenceless) reranker and then do MBR over only the most highly ranked hypotheses, which they also use as the evidence set. \citet{eikema-aziz-2022-sampling} introduce a method, Coarse-to-Fine MBR, that first uses MBR with a cheap-to-compute metric to filter a large hypothesis space to a smaller set, then uses MBR with a better but more expensive to compute metric over the smaller set; they separate evidence and hypothesis sets. \cite{freitag2023epsilon} further investigates sampling strategies for MBR, finding that epsilon sampling \cite{hewitt-etal-2022-truncation} outperforms other strategies in automated and human evaluations.

Another earlier line of work has considered growing \textit{post hoc} the hypothesis set in order to obtain hypotheses with higher expected gain \cite{gonzalez-rubio-etal-2011-minimum, gonzalez-rubio-casacuberta-2013-improving, hoang2021ensembling}.

\subsection{Sampling an evidence set}
\label{sec:sample-evidence}
Comparatively less work has studied strategies for sampling the evidence set. Most recent work has adopted the unbiased sampling strategy of \cite{eikema-aziz-2020-map}, i.e. drawing i.i.d. samples from the model distribution $p(y|x)$ (equation \ref{eq:modeldist}). This strategy is motivated by their observation that unbiased sampling is reasonably reflective of the data distribution, much more so than beam search. However, their approach is incompatible with models trained via label smoothing \cite{Szegedy_2016_CVPR}. \cite{yan2023dcmbr} attempt to remedy this by sampling the evidence set with temperature $\tau < 1$, sharpening the model distribution.

\subsection{What metric do we want to maximize?}
\label{sec:whichmetric}
The gain $G$ (alternatively, error $L$) may be an arbitrary function $\mathcal{Y}_e \times \mathcal{Y}_h \rightarrow \mathbb{R}$. Early work focused on simple, token-level metrics like word error rate and BLEU \cite{kumar-byrne-2004-minimum,ehling-etal-2007-minimum}, but more recent work has explored the use of neural metrics \cite{amrhein-sennrich-2022-identifying,freitag-etal-2022-high}, as well as executing outputs in code generation \cite{shi-etal-2022-natural,Li2022competition}.

Generally, for both neural and non-neural metrics, MBR with metric $G$ as a gain function will yield the largest downstream improvements on $G$ \cite{muller-sennrich-2021-understanding, freitag-etal-2022-high, fernandes-etal-2022-quality}. In other words, if one aims to optimize system performance on metric $M$, one should perform MBR with $M$ as gain. 

However, MBR also inherits the weaknesses and biases of the gain metric used. MBR has been shown to suffer from length and token frequency biases brought on by the metric, i.e. MBR with BLEU prefers shorter sentences \cite{nakov-etal-2012-optimizing, muller-sennrich-2021-understanding}. Similarly, \cite{amrhein-sennrich-2022-identifying} find that MBR over COMET causes higher rates of errors for named entities and numbers due to a lack of sensitivity in the metric. Moreover, MBR is susceptible to overfitting to the metric; \cite{freitag2023epsilon} show that the MBR setting that maximizes the metric is not the one that humans prefer. 

Note that in the most trivial case, where the metric is 
$G(y, y') = \mathbbm{1}[y = y']$,
MBR recovers mode-seeking methods like beam search-- i.e.  MBR under this metric, in expectation, yields the maximum likelihood decoding.

\subsection{What probability distribution should we use to estimate risk?}
\label{sec:whichprob} 
Most MBR decoding methods use the model's score distribution over outputs, $s$, as the (unnormalized) evidence distribution. Alternately, this distribution may be normalized by a temperature (during minimum risk training \cite{smith-eisner-2006-minimum} or decoding \cite{yan2023dcmbr}). 
Some work \cite{, suzgun-etal-2023-follow}
interprets this as a weak proxy for the human or true distribution, arguing that the true objective is to minimize error under the human distribution:
\begin{align*}
    \argmin_{y' \in \mathcal{Y}_h} \mathbb{E}_{y \sim p_\text{human}}[L(y, y')]
\end{align*}
Note that this is not the only reasonable choice of $p(y|x)$; other possible distributions include a distribution over outputs from multiple models (\S \ref{sec:ensemblingismbr}) or the length-penalized distribution over a single model's outputs $p_l(y|x)$ (\S \ref{sec:lengthcorr-exp}).

\section{MBR as a frame for other methods}
\label{sec:itsallmbr}
Self-consistency, range voting, output ensembling, and density estimation can all be viewed through the framing of MBR. This exposes unstated connections between the methods and provides some theoretical backing to the empirical success of these methods. We discuss each in turn.

\subsection{Self-consistency as MBR}
\label{sec:selfconsistency}
Self-consistency \cite{wang2023selfconsistency} is a method for choosing outputs from language models. In self-consistency, the model is prompted 
to generate an explanation and then an answer. 
Multiple outputs $\mathcal{O} = \{y_1, \ldots, y_m\}$ are sampled from the model, the  answers $\mathcal{A} = \{a_1, \ldots, a_m\}$ are extracted $a_i = \mathrm{ans}(y_i)$, and the most frequent answer is returned:
\begin{equation}
    \argmax_a \sum_{i=1}^m \mathds{1}(a_i = a)
\end{equation}

Self-consistency only computes exact match over the \textit{answer}, not the reasoning chain. It is possible to recover MBR from this method by either taking the hypothesis/evidence sets to be the set of resulting answers $\mathcal{Y}_h = \mathcal{Y}_e = \mathcal{A}$  discarding the reasoning chain, or by defining a gain function $G(y, y') = \mathds{1}(\mathrm{ans}(y) = \mathrm{ans}(y'))$  over full outputs $\mathcal{O}$; though notationally different, they are mathematically equivalent.

Thus, self-consistency is a type of MBR decoding in which we approximate the risk with a Monte Carlo estimate (cf. Eq. \ref{eq:mbr-mc}), the answers are sampled from the model (conditioned on the prompt), and the metric is exact match of the ``final answer.''

This framing additionally explains some results from the self-consistency paper. \citet{wang2023selfconsistency} compare the performance of self-consistency across sampling strategies, finding that the best of the strategies they tried are those that are closest to ancestral sampling (nucleus sampling with $p=0.95$ and $\tau=0.7$ without top-k sampling). They also find that self-consistency works better with a sampled output rather than outputs from beam search (their Table 6). 
Through the lens of MBR, this empirical result has a clear theoretical justification: ancestral sampling of evidence sets generally yields the best performance for MBR because this provides an unbiased estimator of the probabilities of the sampled sequences. This also presents an opportunity for improvement: while \citet{wang2023selfconsistency} do not evaluate on ancestral sampling, it is possible that this would outperform their best results.

Self-consistency is a special case of MBR. Proposed extensions to self-consistency have recovered aspects of generalized MBR decoding, including filtering to smaller hypothesis/evidence sets \cite{Fu2022ComplexityBasedPF} and the use of alternative gain metrics \cite{jain2023selfconsistency}.
As a result, the term \textit{self-consistency} has widened in definition from a specific type of MBR to a catch-all for MBR-based decoding methods on large language models.

\subsection{Output Ensembling as MBR}
\label{sec:ensemblingismbr}
Model ensembling techniques that operate on \textit{completed outputs} of models may also be cast in MBR terms. Note that this does not include methods that operate on model weights or partial outputs. Common ensembling methods such as averaging model weights \cite{Izmailov2018AveragingWL} or averaging token-level probabilities \cite{sennrich-etal-2016-improving,manakul-etal-2023-cued} cannot be explicitly formulated as MBR.

The connection to MBR is most straightforward in methods that perform MBR decoding over the outputs of multiple models \citep[\textit{inter alia}]{denero-etal-2010-model, duh-etal-2011-generalized, barzdins-gosko-2016-riga, lee-etal-2022-maximum}. Representative of this family of methods is Post-Ensemble \cite{kobayashi-2018-frustratingly}, which ensembles multiple text generation models $\theta_1,\theta_2,\ldots,\theta_n$ by separately decoding from each model, computing pairwise sentence embedding similarity between all pairs of outputs, and yielding the output with greatest average similarity. Observe that this may be framed as MBR minimizing the expected risk over the mixture distribution
\begin{align*}
p_\text{ensemble}(y|x) = \begin{cases}
    p_{\theta_1}(y|x) & \text{ with probability }\pi_1 \\
    \cdots \\
    p_{\theta_n}(y|x) & \text{ with probability }\pi_n \\
\end{cases}
\end{align*}
where $\sum_{i=1}^n \pi_i = 1$. While $\pi_i$ is usually taken to be uniform over the ensemble, this need not always be the case \cite{duan-etal-2010-mixture}.

Other methods may be viewed as relaxations of MBR decoding. Assemble!~\cite{martinez-lorenzo-etal-2023-amrs} ensembles Abstract Meaning Representation (AMR) graph parsers by computing the pairwise perplexities of each output under \textit{each parser}. 
While this is not precisely MBR, it may be viewed as a variation where the evidence set is \textit{a set of models}, not a set of model outputs.
\begin{align*}
\hat{y} = \argmin_{y' \in \mathcal{Y}_h} \mathbb{E}_{\theta \sim \pi(\cdot)}[L(\theta, y')]
\end{align*}
In this case, the error $L(\theta, y')$ is the perplexity of $y'$ under model $\theta$, i.e. $\exp(-\log p_\theta(y')) = \frac{1}{p_\theta(y')}$, and $\pi(\cdot)$ is the distribution over models.

\subsection{MBR as Density Estimation}
\label{sec:densityest}

Interestingly, Post-Ensemble \cite{kobayashi-2018-frustratingly} (\S\ref{sec:ensemblingismbr}) was not formulated as MBR (and in fact never referred to by name as MBR), but rather as kernel density estimation. Kernel density estimation is a non-parametric method for estimating the probability density function $p$ of an unknown distribution, given samples $(x_1, x_2, \cdots, x_n)$ from that distribution \cite{rosenblatt1956remarks, parzen1962estimation}.
\begin{align}
    \label{eq:densityest}
    \hat{p}(x) = \frac{1}{n} \sum_{i=1}^n K(x, x_i)
\end{align}
Indeed, Equation \ref{eq:densityest} very closely resembles the Monte Carlo estimator of expected loss in Equation \ref{eq:mbr-mc}. This connection allowed \cite{kobayashi-2018-frustratingly} to propose approximation error bounds on MBR, drawing from the density estimation literature.\footnote{We do not reproduce their bounds here; we direct interested readers to the original paper.} 

Note that the kernel function $K(x, x_i)$ is more commonly written as $K(x - x_i)$, or $K(x^T x_i)$ for directional statistics. While this may seem limiting, we can rewrite commonly used MBR metrics in this form; we show this for ROUGE-$n$ as an example. For a sequence $y$, define $T_n(y)$ to be a vector of size $|V|^n$, where $|V|$ is the size of the vocabulary, containing the number of times every possible $n$-gram appears in $y$. Then we can rewrite ROUGE-$n$ as the following:
\begin{align}
    K_\text{R}(T_n(y) - T_n(y')) \nonumber \\    = 1 - \frac{|T_n(y) - T_n(y')|_1}{|T_n(y)|_1 + |T_n(y')|_1}
\end{align}
where $|\cdot|_1$ is the $L1$ norm.

The similarity between density estimation and MBR yields an alternative interpretation of MBR as a mode-seeking search. However, we are not seeking the mode of the model's distribution over outputs, $p(y|x)$, but rather that of a distribution over some features $\phi(y)$ of our output, $p'(\phi(y) | x)$. For instance, in the case of ROUGE-$n$ MBR, 
\begin{align}
    \hat{y} &= \argmax_{y' \in \mathcal{Y}_h} \sum_{y \in \mathcal{Y}_e} K_\text{R}(T_n(y') - T_n(y)) \\
    &\approx \argmax_{y' \in \mathcal{Y}_h} p'(T_n(y') | x)
\end{align}
We posit that this alternative distribution $p'(T_n(y') | x)$ may be better correlated with performance on specific downstream metrics than the original model distribution, potentially adding an additional justification for MBR's effectiveness. 
We hope this may inspire future work investigating the theoretical underpinnings of MBR.

\subsection{Range Voting as MBR}
\label{sec:rangevoting}

Methods that take inspiration from outside of NLP may also be MBR-like; in particular,
some MBR-like algorithms in the literature are formulated from a voting theory perspective where candidate hypotheses are assigned votes based on similarity to some set of voters \cite{wang2023selfconsistency,jain2023selfconsistency,suzgun-etal-2023-follow,hoang2021ensembling}. We show here that range voting \cite{borgeaud-emerson-2020-leveraging}, which broadly encapsulates these proposed voting methods, reduces to MBR.

Range voting describes a family of voting systems in which each voter assigns each candidate a score and the candidate with the greatest total or average score is elected. Observe that the set of candidates $C$ corresponds to the hypothesis set $\mathcal{Y}_h$ and the set of voters $V$ corresponds to the evidence set $\mathcal{Y}_e$. Then, if voter $v$'s score for candidate $c$ is taken to be a gain $G(v, c)$ and each voter is assigned uniform weight, range voting is equivalent to the MBR decision rule in Equation \ref{eq:mbr-mc-gain}:
\begin{equation}
    c_\text{elected} = \argmax_{c \in C} \frac{1}{|V|} \sum_{v \in V} G(v, c)
\end{equation}

Other range-voting methods can similarly be cast as MBR variants.

\section{Design Decisions Impact MBR Performance}
\label{sec:experiments}

Although all the methods in Section~\ref{sec:itsallmbr} are MBR-like, they make very different decisions about the four design choices in our MBR taxonomy. To demonstrate the importance of the method design, we consider empirically two cases where changing design impacts the performance of the method.

\subsection{Experimental Details}
We run MBR experiments for abstractive summarization on CNN/DM \cite{nallapati-etal-2016-abstractive} with a fine-tuned BART-Large\footnote{\texttt{facebook/bart-large-cnn} on HuggingFace \cite{wolf-etal-2020-transformers}} released by the BART authors \cite{lewis-etal-2020-bart} as our base model. In \S \ref{sec:lengthcorr-exp}, we additionally report results for translation on WMT'16 Romanian-English (Ro-En) \cite{bojar-etal-2016-findings} using mBART-50 \cite{liu-etal-2020-multilingual-denoising}.\footnote{\texttt{facebook/mbart-large-50-many-to-many-mmt}} We draw $n_e$ ancestral samples for our evidence set and $n_t$ temperature samples ($\tau = 0.5$ for CNN/DM, $\tau = 0.3$ for WMT'16 Ro-En) for our hypothesis set. We set $n_e = n_t = 30$ in \S \ref{sec:metric-exp} and $n_e = n_t = 50$ in \S \ref{sec:lengthcorr-exp}. Unless otherwise specified, we take ROUGE-1 \cite{lin-2004-rouge} as our gain metric for summarization and BLEU-4 \cite{papineni-etal-2002-bleu}\footnote{We use the implementation from \texttt{sacrebleu} \cite{post-2018-call}  with signature \texttt{nrefs:1|case:mixed|eff:yes|tok:13a|\\smooth:exp|version:2.3.1}} as our gain metric for translation.

\subsection{The MBR metric matters -- but perhaps not as much as the hypothesis set}
\label{sec:metric-exp}
We find that using MBR with the summarization n-gram metric ROUGE-1 \cite{lin-2004-rouge} improves abstractive summarization performance over beam search on CNN/DM, even when evaluating performance with neural metrics; using the general-purpose neural metric BERTScore \cite{zhang2020bertscore} as the MBR metric yields highest BERTScore but smaller gains on non-neural metrics, a finding consistent with past work; and even \textsc{BEER} \cite{stanojevic-simaan-2014-fitting}, a translation metric, works as an MBR metric for this task.

However, prior work using the same dataset and model \cite{wiher-etal-2022-decoding} found that \textsc{BEER} \cite{stanojevic-simaan-2014-fitting} underperforms beam search. This divergence in results is likely due to our different choices in hypothesis set -- \citet{wiher-etal-2022-decoding} use the evidence set plus additional outputs from other decoding methods as hypotheses, while we use temperature samples at $\tau = 0.5$. While reusing the evidence set is more efficient than sampling a separate set of hypotheses, it leads to performance degregation in this case; this further emphasizes the importance of choosing the hypothesis set in MBR. 

\begin{table}
    \centering
    \resizebox{0.5\textwidth}{!}{
    \begin{tabular}{l r r r r}
        \toprule

        Method & R1 & R2 & RL & \textsc{BS} \\
        \midrule 
        Greedy & 43.98 & 20.88 & 30.88 & 88.04 \\
        BS ($k$ = 5) & 43.16 &  20.63 & 30.53 & 87.82 \\
        BS ($k$ = 10) & 42.62 &  20.23 &  30.02 & 87.71 \\
        DBS ($k = g = 5$) & 43.77 & 20.85 & 30.77 & 87.97 \\
        \midrule
        MBR ROUGE-1 & \textbf{46.89} & 22.29 & 32.01 & 88.41 \\
        MBR BEER & 46.31 & \textbf{22.36} & 32.02 & 88.38 \\
        MBR \textsc{BertScore} & 46.04 & 22.09 & \textbf{32.09} & \textbf{88.68} \\
        \bottomrule
    \end{tabular}
    }
    \caption{MBR results on CNN/DM for various gain functions. We additionally test the same non-MBR, (approximate) mode-seeking baselines as \citet{wiher-etal-2022-decoding}. All MBR methods outperform all non-MBR methods tested.}
    \label{tab:mbrsum}
\end{table}

\subsection{Varying the risk distribution: lessons from beam search don't translate to MBR }
\label{sec:lengthcorr-exp}

By nature, autoregressive text generation models suffer from length bias: sequence probability monotonically decreases with increasing length, causing shorter, potentially less informative sequences to be favored by the model distribution \cite{koehn-knowles-2017-six,stahlberg-byrne-2019-nmt}. 
For non-sampling methods such as beam search, the sequence probabilities are generally modified with a length-dependent term when comparing sequences \cite{murray-chiang-2018-correcting, cho-etal-2014-properties}.
Hence, it stands to reason that a length-corrected distribution with these biases alleviated may provide a better estimate of the risk $R(y')$.

Vanilla Monte Carlo MBR (as depicted in Equation \ref{eq:mbr-mc}) yields an estimate of the expected risk under the distribution that our evidence samples are drawn from. To modify the distribution used in our estimate, we turn to \textbf{importance sampling}, a method for estimating the expected value of a quantity under target distribution $p$, given samples from proposal distribution $q$ \cite{kloek1978bayesian}. For a brief tutorial on importance sampling and description of our estimator, see Appendix~\ref{sec:appendix-importance}. 

We take the \emph{score} of a sequence to be the log probability: 
We then experiment with two of the strategies described in \cite{murray-chiang-2018-correcting} for constructing the length corrected score $s_l(y|x)$:
\begin{enumerate}[(a), nosep]
    \item \textbf{Length normalization}: The model distribution is smoothed with temperature $T^\beta$, where $T$ is the sequence length and $\beta$ is the length penalty, a hyperparameter. A larger $\beta$ more heavily prioritizes longer sequences.
        \begin{align}
        \label{eq:length-norm}
            s_l(y|x) = s(y|x) / T^\beta
        \end{align}
    \item \textbf{Length reward} \cite{he2016improved}: A fixed reward $\gamma$ is added to the score per token generated.
        \begin{align}
        \label{eq:length-reward}
            s_l(y|x) = s(y|x) + \gamma T
        \end{align}
\end{enumerate}
The length-corrected distribution is then $p_l(y|x) \propto \exp s_l(y|x)$. We apply \textbf{normalized importance sampling} \cite{rubinstein2016simulation} to estimate the risk under the length corrected distribution, i.e. $R(y') = \mathbb{E}_{y \sim p_l}[L(y,y')]$, given samples drawn from the model distribution $p(y|x)$.

We compare our MBR results against beam search both with and without length normalization. We use the models' default values for length penalty ($\beta = 2$ for BART, $\beta = 1$ for mBART).

\begin{table}
    \centering
    \resizebox{0.5\textwidth}{!}{
    \begin{tabular}{l r r r r r}
        \toprule

        Method & R1 & R2 & RL & \textsc{BS} & LR \\
        \midrule 
        Beam search, no correction &  43.88 & 20.96 & 30.77 & 87.79 & \textbf{108.00} \\
        Beam search & 43.95 & 21.00 & 30.84 & 87.81 & 114.39 \\
        \midrule
        MBR, No correction & \textbf{47.70} & \textbf{23.00} & \textbf{32.54} & \textbf{88.50} & 111.64 \\
        MBR, Length norm, $\beta = 0.5$ & 44.29 & 19.95 & 29.99 & 88.03 & 110.75 \\
        MBR, Length norm, $\beta = 1.0$ & 44.29 &  19.98 &  30.0 & 88.03 & 110.77 \\
        MBR, Length reward, $\gamma = 0.5$ & 47.60 &  22.93 &  32.48 & 88.48 & 112.52 \\
        MBR, Length reward, $\gamma = 1.0$ & 47.41 & 22.72 & 32.25 & 88.43 & 112.50 \\
        \bottomrule
    \end{tabular}
    }
    \caption{MBR results for various length correction schemes on CNN/DM. We report ROUGE-1, ROUGE-2, ROUGE-L, \textsc{BertScore}, and length ratio, respectively.}
    \label{tab:lengthcorrect-cnndm}
\end{table}

\begin{table}
    \centering
    \resizebox{0.5\textwidth}{!}{
    \begin{tabular}{l r r r r r}
        \toprule

        Method & BLEU & chrF & \small{BLEURT} & BS & LR\\
        \midrule 
        Beam search, no correction & 33.21 & 59.81 & 65.50 & 94.95 & 99.37 \\
        Beam search & 33.06 & \textbf{60.05} & \textbf{65.60} & \textbf{94.96} & 101.58 \\
        \midrule
        MBR, No correction & \textbf{33.56} & \textbf{60.00} & 65.53 & \textbf{94.96} & \textbf{100.04} \\
        MBR, Length norm, $\beta = 0.5$ & 31.14 & 58.53 & 64.70 & 94.71 & 102.82 \\
        MBR, Length norm, $\beta = 1.0$ & 31.09 & 58.51 & 64.68 & 94.71 & 102.60 \\
        MBR, Length reward, $\gamma = 0.5$ & 32.09 & 59.63 & 65.19 & 94.82 & 105.00 \\
        MBR, Length reward, $\gamma = 1.0$ & 31.29 & 59.17 & 64.91 & 94.73 & 105.63 \\
        \bottomrule
    \end{tabular}
    }
    \caption{MBR results for various length correction schemes on WMT'16 Romanian-English. We report BLEU, chrF, BLEURT, \textsc{BertScore}, and length ratio, respectively. We use the chrF \cite{popovic-2015-chrf} implementation from \texttt{sacrebleu}. We use the smaller \texttt{BLEURT-20-D6} checkpoint for efficiency \cite{sellam-etal-2020-bleurt,pu-etal-2021-learning}.}
    \label{tab:lengthcorrect-wmt}
\end{table}

Our results are Tables \ref{tab:lengthcorrect-cnndm} and \ref{tab:lengthcorrect-wmt}. In line with past work, we find that beam search generally benefits from incorporating a length penalty. However, we find that length-corrected MBR underperforms vanilla MBR. 
This may be due to a gap between the sampling and length-correction distibutions, leading to a high-variance estimator of risk. 

However, our results are also emblematic of a wider trend among minimum-risk techniques. Past work has found that models trained with Minimum Error Rate Training \cite{och-2003-minimum,shen-etal-2016-minimum}, an error-aware training method, do not require length correction in beam search \cite{neubig-2016-lexicons}. Similarly, we find that MBR without length correction generates outputs relatively close in length to the references, more so than length-normalized beam search.
 This suggests that MBR may be to some extent immune from length biases, when they are not introduced by the MBR metric \cite{muller-sennrich-2021-understanding}.

\section{MBR applications in NLP}
\label{sec:mbrnlp}
The use of minimum Bayes risk decoding in NLP predates these MBR-like methods; MBR has been applied by name in NLP since the 1990s.
\paragraph{Historical context}
\label{sec:historical-use}
 Minimum Bayes Risk decoding has roots in Bayesian decision theory, a field of study that dates as far back as the Age of Enlightenment \cite{bernoulli1738specimen,Parmigiani2001}. Central to Bayesian decision theory is the principle of risk minimization: in the face of uncertainty, an optimal decision maker should choose the option that minimizes the amount of error they can expect to suffer -- or, in other terms, maximizes the amount of utility they can expect to enjoy \cite{degroot-1970,bickel-doksum-1977}. This is precisely the intuition encoded in MBR (i.e. Equation \ref{eq:mbr-risk}).
 
\paragraph{Adoption in NLP} MBR was adopted by the speech and NLP communities in the 1990s and early 2000s, finding applications in syntactical parsing \cite{goodman-1996-efficient, simaan-2003-maximizing}, automatic speech recognition \cite{stolcke1997, Goel2000}, and statistical machine translation \cite{kumar-byrne-2004-minimum, tromble-etal-2008-lattice, kumar-etal-2009-efficient}. Many NLP tasks during this time relied upon graph structures as inductive biases (i.e. parse trees or translation lattices/hypergraphs). As such, early MBR works often used these graphical models as hypothesis and evidence spaces. Work on lattice MBR \cite{tromble-etal-2008-lattice}, for instance, treated the set of all hypotheses encoded in a word lattice, of which there are exponentially many, as both evidence and hypothesis sets. This is in contrast to most later MBR work, which operates on a relatively small list of text outputs obtained from a neural model. As a result, early work relied on rather involved dynamic programming algorithms for exact MBR decoding and were restricted to token-factorizable metrics such as BLEU and edit distance. Later work additionally demonstrated the efficacy of MBR for question answering \cite{duan-2013-minimum} and for joining statistical and neural approaches to translation \cite{stahlberg-etal-2017-neural}.

\begin{figure}
    \centering
    \includegraphics[width=0.48\textwidth]{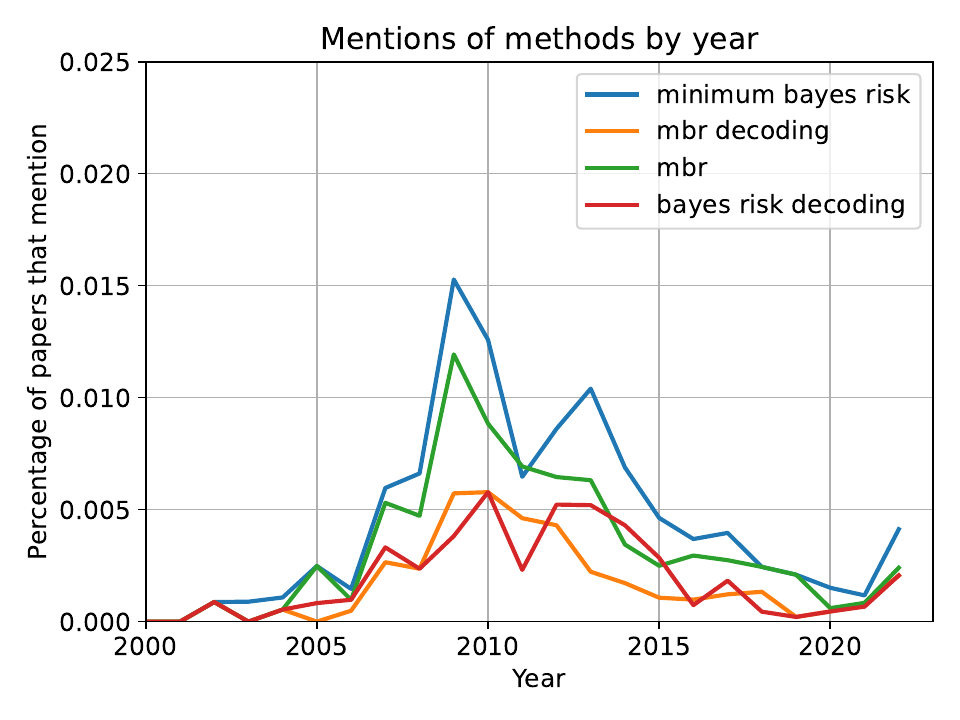}
    \caption{The use of MBR (by name) peaked in the mid-2010s. This graph shows the percentage of ACL Anthology papers that mention several MBR-related phrases by year, from 2000 to 2022.}
    \label{fig:mbr-usage}
\end{figure}

\label{sec:recent-use}
 
\paragraph{Recent usage} In an effort to move past beam search, which has well-known pathologies \cite{stahlberg-byrne-2019-nmt}, MBR has in recent years resurfaced as a decision rule for text-generation models \cite{eikema-aziz-2020-map}. As discussed earlier in \S \ref{sec:best-practices}, several lines of work have sprung up investigating the properties of MBR in modern neural text generation setups. Notably, however, most of these works have focused on applications of the method to neural machine translation, with only a few very recent works studying its applications in other text generation tasks \cite{shi-etal-2022-natural,wiher-etal-2022-decoding, suzgun-etal-2023-follow}.

Outside of these areas, the method has largely been applied in shared task papers (e.g. \citet{manakul-etal-2023-cued,yan-etal-2022-cmus,barzdins-gosko-2016-riga}), as it provides a reliable boost in performance.  The fraction of papers in the ACL Anthology that reference MBR (at least by this name) has declined from its peak around 2009 (Figure \ref{fig:mbr-usage}).

\section{Conclusion}

Minimum Bayes Risk decoding has declined in popularity, but the underlying concept of sampling a set from a distribution and choosing an output to minimize risk according to that set has remained. This concept now takes many surface forms-- from self-consistency to range voting to output ensembles-- and current research in these areas rarely draws connections to MBR.
While rediscovery is a key part of science, so is recontextualizing new methods within a broader research narrative. This can often reveal new insights or cast findings in a different light.
For instance, the empirical benefits of self-consistency can be justified through an MBR framing; work on extensions to self-consistency has rediscovered other properties of MBR; and work on ensembling has raised questions about how to weight mixtures of models that can be reasoned about within the framework of noisy estimates of global probability distributions.

The adoption of newer terms for MBR-like methods may be a type of terminology drift. Related phenomena have been studied in the philosophy of science literature, including pressures to coin new terms \cite{neologisms-recycle, merton-priorities}, potential negative consequences of divergent terminology \cite{psychologists-airmen, oil-translation-gaps}, and decreased citation of older methods in NLP \cite{singh-etal-2023-forgotten}.  For a more involved discussion of the literature on term coining and possible connections, see Appendix~\ref{sec:appendix-philsci}.

Language is not static, so some degree of terminology drift in scientific literature is unavoidable. However, recognizing the connections between modern techniques and older work is crucial to understanding why such methods are effective. 
We must not forget the lessons of the past as we search for the methods of the future.

\section*{Acknowledgments}
We would like to thank Jason Eisner and Patrick Fernandes for useful early discussions about this work, and Saujas Vaduguru, Daniel Fried, and Shuyan Zhou for feedback on this draft.

This work was supported in part by grants from the Singapore Defence Science and Technology Agency, 3M — M*Modal, the Air Force Research Laboratory (AFRL), and the National Science Foundation Graduate
Research Fellowship Program under Grant No. DGE2140739. Any opinions, findings, and conclusions or recommendations expressed in this material are those of the author(s) and do not necessarily reflect the views of the sponsors. 

\bibliography{anthology,custom}
\bibliographystyle{acl_natbib}

\onecolumn
\newpage 
\appendix
\twocolumn 

\section{More details on importance sampling for MBR}
\label{sec:appendix-importance}
We present in this section the normalized importance sampling estimator of risk used in our experiments in \S \ref{sec:lengthcorr-exp}.

The core insight of importance sampling is that we can rewrite the expected value of a random variable $f(x)$ under target distribution $p$ as another expectation under some proposal distribution $q$:
\begin{align*}
    \mathbb{E}_p[f(x)] &= \sum_{x} f(x) p(x) \\
    &= \sum_{x} f(x) \frac{p(x)}{q(x)} q(x) \\
    &= \mathbb{E}_q\bigg[f(x) \frac{p(x)}{q(x)}\bigg]
\end{align*}

Importance sampling can be particularly useful when sampling from the proposal distribution is easy, but sampling from the target distribution is costly or intractable; this is indeed the case for MBR, as sampling from the length-corrected distribution $p_l(y|x)$ requires computation of its partition function, which has exponential complexity.

Hence, for MBR, if we draw evidence samples $\mathcal{Y}_e$ according to model distribution $p(y|x)$ but wish to compute the risk under some length-corrected distribution $p_l(y|x)$, we may compute
\begin{align*}
    R(y') &= \mathbb{E}_{y \sim p_l}[L(y, y')] \\
    &= \mathbb{E}_{y \sim p}\bigg[L(y, y') \frac{p_l(y|x)}{p(y|x)}\bigg] \\
    &= \sum_{y \in \mathcal{Y}_e} L(y, y') \frac{p_l(y|x)}{p(y|x)} \\
    &= \sum_{y \in \mathcal{Y}_e} L(y, y') w(y)
\end{align*}
where we let $w(y) = p_l(y|x) / p(y|x)$, commonly referred to as the importance weight.

Note, however, that importance sampling requires us to be able to exactly compute the probabilities $p(y|x)$ and $p_l(y|x)$; while the former can be computed efficiently (Equation \ref{eq:modeldist}), the latter is intractable, again because it requires the partition function. What we can efficiently compute is the 
unnormalized probability 
$\Tilde{p_l}(y|x) = \exp s_l(y|x)$, where $s_l$ is the length-corrected score given by either Equation \ref{eq:length-norm} or \ref{eq:length-reward}.

Fortunately, we can use \textbf{normalized importance sampling} to obtain a consistent estimator of the risk by adjusting importance weights \cite{rubinstein2016simulation}:
\begin{align}
    R(y') &= \mathbb{E}_{y \sim p_l}[L(y, y')] \\
    &= \frac{ \mathbb{E}_{y \sim p}[L(y, y') \Tilde{w}(y)] }{ \mathbb{E}_{y \sim p}[\Tilde{w}(y)] } \\
    &= \sum_{y \in \mathcal{Y}_e} L(y, y') \cdot \frac{ \Tilde{w}(y) }{ \sum_{y \in \mathcal{Y}_e} \Tilde{w}(y) }
\end{align}
where $\Tilde{w}(y) = \Tilde{p_l}(y|x) / p(y|x)$. As it is the ratio of two estimates, the normalized importance sampling estimator is \textit{biased} for finite sample sizes.
\newpage
\section{Contextualizing this work within philosophy of science}
\label{sec:appendix-philsci}

In this section, we contextualize our work in the broader framings of meta-analysis of scientific research. 
\paragraph{Patterns of citation in NLP} Several factors have been shown to correlate with citation rate in NLP, including author geographic location \cite{rungta-etal-2022-geographic}, author gender \cite{mohammad-2020-gender}, and publication date \cite{bollmann-elliott-2020-forgetting,singh-etal-2023-forgotten}. \citet{bollmann-elliott-2020-forgetting} conduct a bibliometric anaylsis of the ACL Anthology, finding that the mean age of papers cited decreased significantly from 2010 to 2019. \citet{singh-etal-2023-forgotten} expand this analysis to the full anthology, finding that, while citations of older papers rose briefly in the mid-2010s, it has since declined, with 2021 marking a historic low for the percentage of citations that went to older papers\footnote{They define an ``older paper'' as one that is more than 10 years older than the paper that is citing it.}. They term this \textit{citational amnesia} and discuss several possible reasons for the result, including the shift to neural methods and the rise of new areas of NLP.

Our work raises another potential explanation: some citational amnesia is due to \textit{terminology drift} over time, as old methods begin to be referred to by newer names. 

\paragraph{Term coining in science} Work in science and technology studies has examined the broader phenomenon of term coining in science. 
\citet{neologisms-recycle} argues that neologisms emerge more frequently in fields that prize novelty and see science as fundamentally about leaps of discovery, and fields that are perceived as synthesizing findings from multiple fields are most likely to recycle terms from other disciplines. She cites computer science as an example of a field where most new terms of art emerge from recycling common words, often those that draw a metaphor to some basic physical or human concept; this is reflected in the adoption of the humanizing ``self-consistency'' and the political-science-inspired ``range voting'' in decoding. \citet{evocative-publicity} suggests that evocative, metaphor-laden names are more likely to emerge as a scientific field grows more public-facing and in times where many new terms are being coined; both of these descriptors apply to modern NLP. While several works in linguistics and STS have considered the coining of new terms for new phenomena, relatively little work has focused on the divergence of terminology for previously observed phenomena. 

The consequences of divergent or distinct terminology have also been studied, with differences in terminology across fields blamed for slow adaptation of research to practical applications (e.g. in studying visual distortions during plane takeoff \cite{psychologists-airmen}). Borrowing terminology from another language (often Latin or Greek) or from another field has been described as a method to build common ground between researchers \cite{oil-translation-gaps} and as a possibly concerning pressure against developing language-specific scientific terminology in lower-resourced languages \cite{danish-domain-loss}. However, most work on lexical divides in science has focused on divides across language or field rather than divides across time in the same field.

\end{document}